\pdfoutput=1
\documentclass[11pt]{article}

\usepackage[final]{coling}

\usepackage{times}
\usepackage{latexsym}

\usepackage[T1]{fontenc}

\usepackage[utf8]{inputenc}

\usepackage{microtype}

\usepackage{inconsolata}

\usepackage{graphicx}

\usepackage{chngcntr}
\counterwithout{figure}{section}
\usepackage[utf8]{inputenc} 
\usepackage{amsmath}
\usepackage[T1]{fontenc}    
\usepackage{hyperref}       
\usepackage{url}            
\usepackage{booktabs}       
\usepackage{amsfonts}       
\usepackage{nicefrac}       
\usepackage{microtype}      
\usepackage{lipsum}
\usepackage{xcolor}
\usepackage{graphicx}
\usepackage{pifont}
\graphicspath{ {./images/} }
\newcommand{\xmark}{\ding{55}}%

\title{Arabic Stable LM: Adapting Stable LM 2 1.6B to Arabic}

\author{Zaid Alyafeai \quad Michael Pieler \quad Hannah Teufel \quad Jonathan Tow \\ Marco Bellagente \quad Duy Phung \quad Nikhil Pinnaparaju \quad Reshinth Adithyan \\
Paulo Rocha \quad Maksym Zhuravinskyi \quad Carlos Riquelme
\\
\\
\textbf{Stability AI Language Team}\thanks{Correspondence to: michael@stability.ai}
}

\begin{document}
\maketitle
\begin{abstract}
Large Language Models (LLMs) have shown impressive results in multiple domains of natural language processing (NLP) but are mainly focused on the English language. Recently, more LLMs have incorporated a larger proportion of multilingual text to represent low-resource languages. In Arabic NLP, several Arabic-centric LLMs have shown remarkable results on multiple benchmarks in the past two years. However, most Arabic LLMs have more than 7 billion parameters, which increases their hardware requirements and inference latency, when compared to smaller LLMs. This paper introduces Arabic Stable LM 1.6B in a base\footnote{\url{https://huggingface.co/stabilityai/ar-stablelm-2-base}} and chat\footnote{\url{https://huggingface.co/stabilityai/ar-stablelm-2-chat}} version as a small but powerful Arabic-centric LLM. Our Arabic Stable LM 1.6B chat model achieves impressive results on several benchmarks beating multiple models with up to 8x the parameters. In addition, we show the benefit of mixing in synthetic instruction tuning data by augmenting our fine-tuning data with a large synthetic dialogue dataset. 
\end{abstract}

\section{Introduction}
Large language models (LLMs) achieved impressive results on multiple zero-shot tasks, where they not only show outstanding results in English but also in other languages \cite{ustun2024aya}. In the past few years, there have been many attempts to create Arabic-centric LLMs. Some of those setups fine-tuned English-focused LLMs, like Llama 2 \cite{touvron2023llama} for AceGPT, \cite{huang2023acegpt}, or were trained from scratch like Jais \cite{sengupta2023jais}. Most Arabic-centric LLMs published on HuggingFace\footnote{\url{https://huggingface.co/models}} have at least 7 billion (B) parameters. This raises the question of whether we can create a smaller but more capable Arabic-centric LLM that can compete with those larger models.

This paper introduces Arabic Stable LM 1.6B, an extended version of Stable LM 2 1.6B \cite{bellagente2024stable} fine-tuned on more than 100 billion Arabic text tokens. We show through extensive evaluations that our fine-tuned models can achieve results on par with models using up to 8x the parameters. Our contributions are summarized as follows:
\begin{enumerate}
    \item We release Arabic Stable LM 1.6B, a state-of-the-art Arabic-centric LLM in its size range when evaluated on cultural alignment and natural language understanding benchmarks. In particular, we show that our Arabic Stable LM 1.6B chat model achieves better results on multiple benchmarks in cloze format (CF), i.e., Arabic cultural alignment and MMLU, when compared to models 8x larger.
    \item We create an Arabic instruction-tuning dataset using LLM-based text rephrasing to train our Arabic Stable LM 1.6B chat model. We use this dataset for fine-tuning and could show that it improves our base model on multiple benchmarks. 
\end{enumerate}
This paper is organized as follows: \textbf{Section \ref{sec:lr}} introduces related work. \textbf{Section \ref{sec:pre}} outlines the used pre-training data and investigates different cleaning procedures for it. \textbf{Section \ref{sec:it}} covers the instruction tuning setup including the various datasets to train the chat models. In \textbf{Sections \ref{sec:eval}} and \textbf{\ref{sec:results}}, we discuss the evaluation setups and results of different benchmarks in detail.

\section{Related Work}
\label{sec:lr}
Since the release of GPT-2 \cite{radford2019language} with its outstanding zero-shot capabilities, there have been many efforts to extend such LLMs to other languages.

For the Arabic language, AraGPT-2 \cite{antoun2020aragpt2} was the first capable model for such use cases. AraGPT-2 was released in four sizes, ranging from 135 million (M) to 1.46 billion (B) parameters. These models were trained on a dataset with approximately 8.8B Arabic words. AraGPT-2 showed impressive Arabic generation results but lacked zero-shot capabilities on downstream tasks, which can be linked to the small pre-training dataset. 
Besides the above work that uses decoder-only models, there have been several attempts to use encoder-decoder models, like T5 \cite{raffel2020exploring}, i.e., AraT5 \cite{nagoudi2021arat5}, AT5B \cite{ghaddar2022revisiting}, and AraMUS \cite{alghamdi2023aramus}.

Recently, larger LLMs have been published for Arabic. Huang et al. \cite{huang2023acegpt} released AceGPT with 7B and 13B parameters in a base and chat version. For those LLMs, they used the Llama 2 7B and Llama 2 13B models and fine-tuned them on 30B and 10B tokens, respectively. The models were further fine-tuned on instruction-tuning data consisting of more than 360K samples. The chat models showed impressive results on multiple Arabic benchmarks, especially in capturing the cultural nuances of the Arabic region. The authors extended their approach by releasing a 1.5 version of their models \cite{zhusecond} by progressively growing the vocabulary size of the Llama 2 base models. Similarly, Sengupta et al. \cite{sengupta2023jais} released the Jais model series, which consists of a 13B and a 30B parameter model trained from scratch on 72B Arabic tokens. In addition, the models were fine-tuned on 3.6M instruction-tuning samples. At the time of their release, the 30B chat model achieved state-of-the-art results on multiple Arabic benchmarks. Furthermore, new Jais models were released that cover a broader model size range from 590M to 30B parameters \cite{jaisfamilymodelcard}. In addition, there were additional attempts to instruct-tune English-centric LLMs on Arabic data. GemmAr-7B \cite{chouikhi2024llamar} was created by fine-tuning Gemma 7B \cite{team2024gemma} on instruction tuning data with around 500k samples. GemmAr-7B outperforms other Arabic LLMs in its size range. Similarly, SILMA \cite{silma_01_2024} is a 9B instruct model built on top of Gemma 2, which achieved remarkable results on multiple Arabic benchmarks. 

Some closed models have been pre-trained on larger Arabic text datasets. The Noor project \cite{lakim2022holistic} fine-tuned a series of models ranging from 1.5B to 13B parameters on 150B Arabic tokens. Similarly, ALLaM \cite{bari2024allam} is a series of models ranging from 7B to 70B parameters that showed impressive results on multiple Arabic and English benchmarks.

\section{Pre-training}
\label{sec:pre}
In this section, we first discuss the concept of tokenization and how it affects Arabic text in particular. Then, we outline our cleaning procedure to preprocess our Arabic data to pre-train our base models. We also discuss our hyperparameter choices for pre-training and experiment with two different learning rate schedulers.

\subsection{Tokenization}
The Stable LM 2 1.6B model was trained on multilingual data, including English, German, French, Italian, Dutch, Spanish, and Portuguese. Our initial tests showed that the model has limited generation capabilities in the Arabic domain (see Appendix \ref{app:zero}). In Table \ref{tab:tok_size}, we compare the tokenizer vocabulary size of different models from the literature, which spans a wide range from a few ten to hundreds of thousands of tokens.

\begin{table}[!htp]
\centering
\caption{Comparison of the vocabulary size of some LLMs in various sizes.}
\label{tab:tok_size}
\begin{tabular}{lr}
\hline
\textbf{Model}          & \textbf{Vocab Size} \\ \hline 
AceGPT-7B          & 32,000     \\ 
jais-13b           & 84,992     \\ 
stablelm-2-1\_6b   & 100,289    \\ 
mt0-xl             & 250,100    \\ 
bloom-7b1          & 250,680    \\ 
c4ai-command-r-v01 & 255,000    \\ \hline
\end{tabular}
\end{table}

\begin{figure}[!htp]
    \centering
    \includegraphics[width=\linewidth]{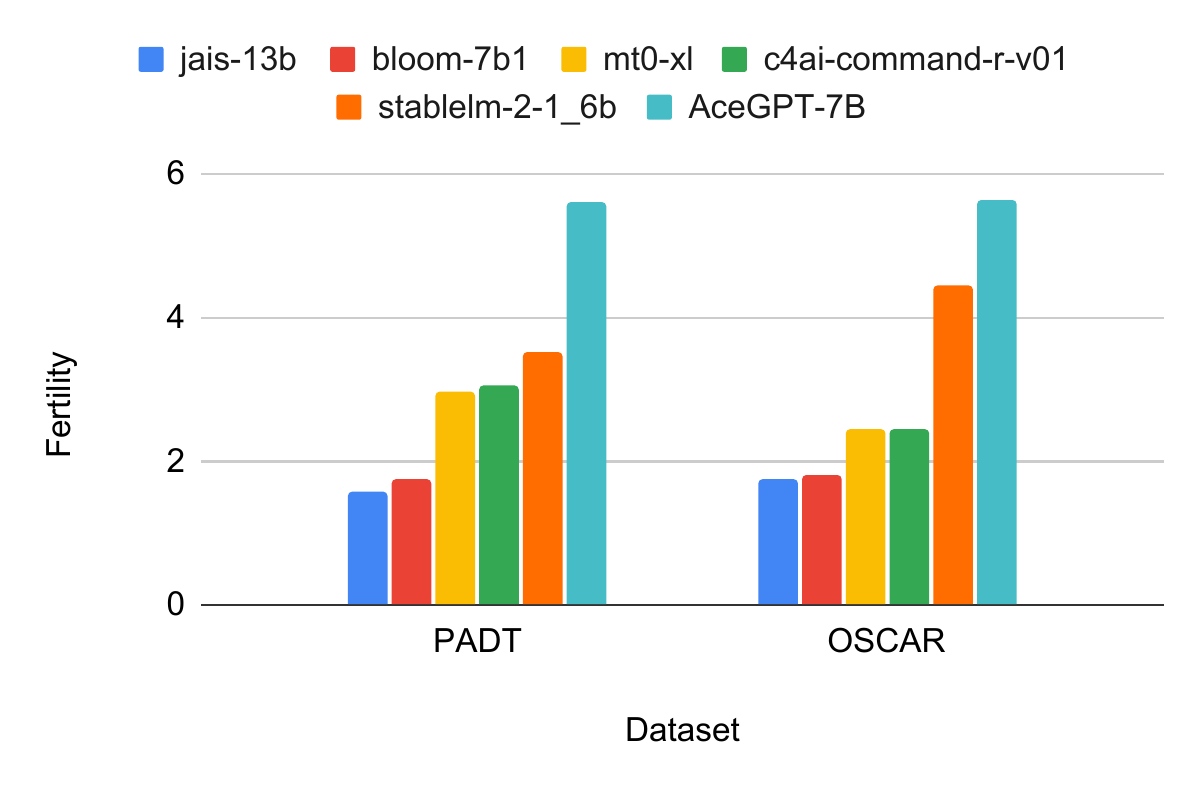}
    \caption{Fertility scores of multiple tokenizers on the PADT and OSCAR datasets.}
    \label{fig:fertility}
\end{figure}

We use the fertility score to measure the "overtokenization", which is defined by the ratio of the sum of (sub-)word tokens divided by the sum of total words \cite{sengupta2023jais}. A higher fertility score results in a higher token count per word, i.e., a word is tokenized with multiple sub-word tokens, where one token for a word would be the lower limit and a single-character tokenization would be the upper limit. In Figure \ref{fig:fertility}, we compare the fertility scores of multiple models from the literature. We use two types of datasets to evaluate the fertility score: The Arabic unshuffled and deduplicated version of OSCAR \cite{2022arXiv220106642A} and the Prague Arabic Dependency Treebank (PADT) dataset \cite{hajic2004prague}. As indicated in Figure \ref{fig:fertility}, we noticed that Jais attains the lowest fertility rate. This is expected since the tokenizer was trained from scratch on Arabic data resulting in the highest compression. In contrast, AceGPT has the highest fertility rate because it is based on the Llama 2 tokenizer with a vocabulary size of 32k, which mostly contains English (sub)-words. The Stable LM 2 1.6B tokenizer achieves the second-highest fertility after the AceGPT models. The tokenizer is based on Arcade100k which extends OpenAI’s tiktoken tokenizer by adding special tokens for code and digit split \cite{li2023starcoder,bai2023qwen,liu2023goat}.

\subsection{Datasets}
For pre-training, we use the English training mixture from  \cite{bellagente2024stable} and combine it with three additional Arabic datasets that include text from the web, news, and books. As shown by \cite{sengupta2023jais}, adding English to the training mixture improved the results for Arabic. Below is the detailed list of the Arabic datasets used in the pre-training mixture:

\begin{enumerate}
    \item \textbf{CulturaX} is a multilingual dataset in 167 languages \cite{nguyen2023culturax}. The Arabic subset contains around 74K documents, which results in more than 158B tokens\footnote{Tokens are evaluated using the StableLM tokenizer.}. This dataset is a combination of the mC4 \cite{2019t5} and OSCAR \cite{suarez2019asynchronous} datasets. 
    \item \textbf{SANAD} is an Arabic news article dataset \cite{einea2019sanad} that contains around 135k documents resulting in 145M tokens. The dataset contains multiple subjects, e.g., culture, finance, medicine, and politics.
    \item \textbf{Arabic E-book corpus} is a freely available collection of 1,745 books consisting of around 280M tokens published between 2008 and 2024 by the Hindawi Foundation \cite{Hallberg2024}. The books cover various genres, including non-fiction, novels, children's literature, poetry, and plays. 
\end{enumerate}

\begin{table}[!htp]
\centering

\caption{Token count in billions (B) for the English and Arabic pre-training datasets. The ratio of English to Arabic tokens is approximately 5:1, and the sampling percentage ratio is set to approximately 4.6:1. The token count is also used to calculate the sampling weights.}
\label{tab:token_count}
{\small
\tabcolsep5pt 
\begin{tabular}{lccc}
\hline
& \textbf{Tokens (B)} & \textbf{Token (\%)} & \textbf{Sampling (\%)} \\ \hline 
\textbf{English} & 619 & 84 & 18  \\ 
\textbf{Arabic} & 115 & 16 & 82 \\ \hline
\textbf{Total} & 734 & 100 & 100  \\ \hline
\end{tabular}
}
\end{table}

 \subsection{Cleaning}
 For cleaning the Arabic datasets, we used the datatrove library\footnote{\url{https://github.com/huggingface/datatrove}} and applied different filtering and cleaning steps to optimize our base data for the pre-training:
\begin{itemize}
    \item \textbf{Safety filtering} We filter the documents based on their URL and content. For Arabic content, we create a list of around 400 phrases that we consider unsafe. Any document that contains at least 3 of such phrases is removed. We remove any document that doesn't contain a source URL. This safety filtering is only applied to the CulturaX data. 
    \item \textbf{Ads filtering} We noticed that CulturaX contains many documents with advertisements. To filter out such documents, we created a list of 12 phrases and removed documents if they contained more than 5. 
    \item \textbf{Line filtering} We filter documents that contain less than 4 lines. We also filter out documents that have more than 50\% lines with less than 3 words. 
    \item \textbf{Character-based filtering} We filter documents that contain less than 95\% of permissible characters, which we define as English and Arabic alphanumeric and special characters. 
    \item \textbf{Gopher filtering} We adapt the Gopher cleaning setup \cite{rae2021scaling} to Arabic by re-calculating the used statistics, e.g., average word length, number of stop words, etc. The Gropher quality filtering contains multiple criteria like filtering based on document length, minimal and maximal word length, the number of special characters, and the percentage of alphabetic characters. We also add an additional step to filter by the number of stop words and punctuation characters. 
    \item \textbf{Character cleaning} We re-map all Arabic characters that have different Unicode characters to the same Unicode character. 
    \item \textbf{Document cleaning} Many documents in our training data contain a title and date at the beginning of each document. We identify such documents and remove this information using regex rules. This mainly affected the CulturaX subset.
\end{itemize}
The total number of documents and tokens before and after cleaning is shown in Table \ref{tab:cleaning_stats}. After applying the cleaning pipeline to the Arabic text we obtained 114B tokens. Because the Stable LM 2 tokenizer has a much higher fertility rate for Arabic text, we sample 4.6 times more Arabic data to remove the fertility bias. The total number of tokens for Arabic and English are shown in Table \ref{tab:token_count}.

\begin{table*}[!htp]
\caption{Total number of tokens and documents before and after cleaning. After cleaning, we drop all the documents that don't meet the filtration criteria. In addition, We show the percentages of documents and tokens that are kept for pre-training. B is billions, M is millions, and K is thousands.}
\label{tab:cleaning_stats}
\centering
{\small
\tabcolsep 10pt 
\begin{tabular}{lcccc}
\hline
                  & \multicolumn{2}{c}{\textbf{Before cleaning}}                                           & \multicolumn{2}{c}{\textbf{After cleaning}}      \\ \hline
\textbf{Dataset}  & \textbf{Tokens} & \textbf{Documents} & \textbf{Tokens} & \textbf{Documents}  \\ \hline
\textbf{CulturaX} & 158.1 B & 74.0 M                              & 114.1 B (72 \%) & 49.9 M (67 \%)   \\ 
\textbf{SANAD}    & 145.1 M     & 134.5 K                                & 140.7 M (95 \%)     & 128.7 K  (96 \%)         \\ 
\textbf{E-Book}   & 280.3 M      & 1.7 K                                   & 171.0 M (61 \%)     & 1.4 K (79 \%)            \\  \hline
\textbf{Total}    &158.6 B & 74.2 M                              & 114.5 B (72 \%) & 50.0 M (68 \%)                    \\ \hline
\end{tabular}
}
\end{table*}

\subsection{Hyperparameters}
\begin{figure}[!htp]
    \centering
    \includegraphics[width=\linewidth]{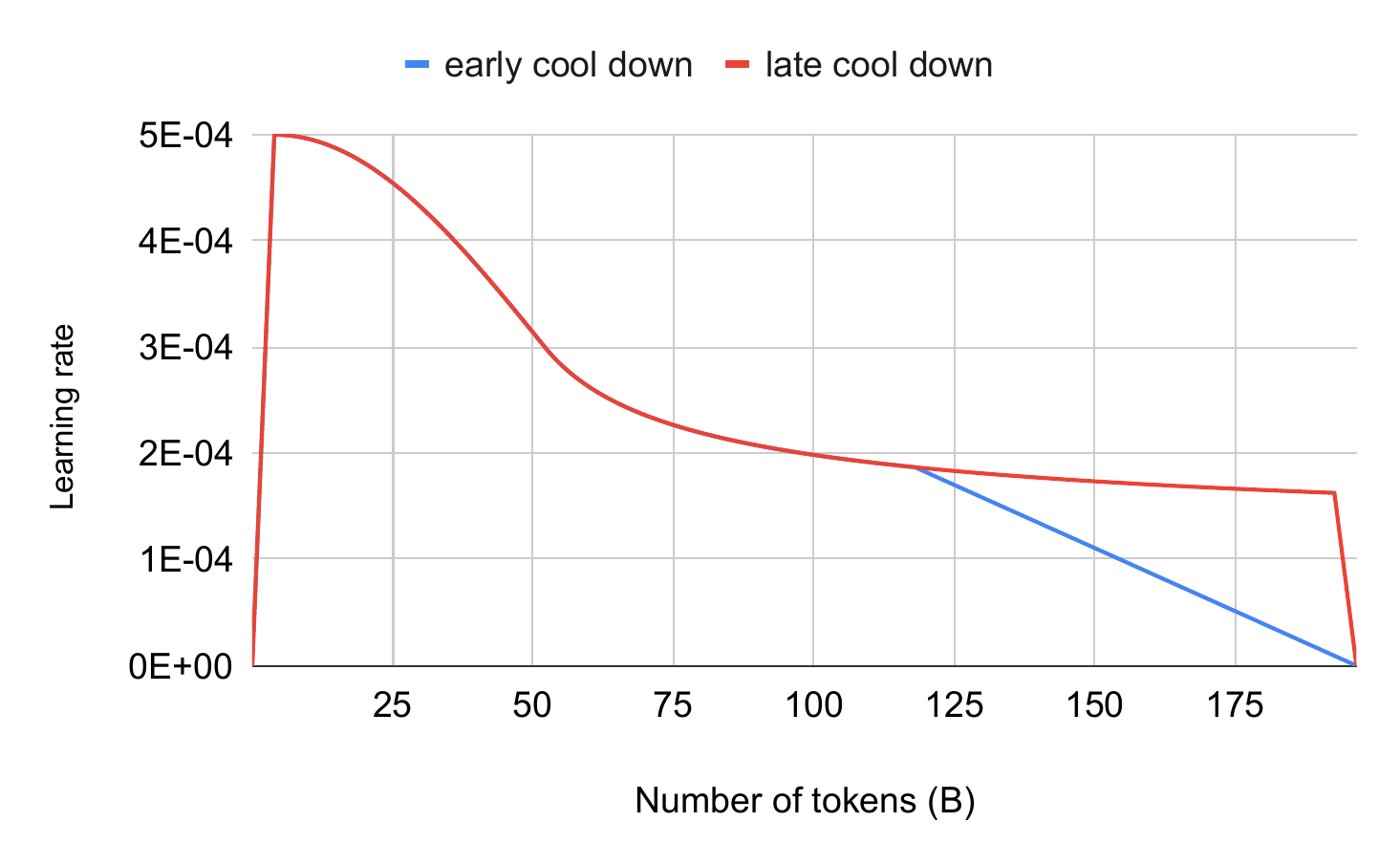}
    \caption{Different learning rate schedulers. The late cool down setup (red) consists of a warm up, cosine and inverse square root, and late cool down phase. The early cool down setup (blue) consists of a warm up, cosine and inverse square root, and early cool down phase.}
    \label{fig:lr}
\end{figure}

We fine-tune the Stable LM 2 1.6B checkpoint\footnote{\url{https://huggingface.co/stabilityai/stablelm-2-1_6b}} for 500k steps in total. This consists of 10k warm-up steps, up to 300k steps with a cosine and inverse square-root learning scheduler, and a cool down using a linear learning rate for 200k steps. The maximum learning rate is set to $5 \times 10^{-4}$ and the minimum learning rate to $2.5 \times 10^{-6}$. The rest of the parameters are set as in the original paper \cite{bellagente2024stable} (see Appendix \ref{app:parameters}). We use two nodes for the training, each with 8 H100 GPUs with a micro-batch size of 6 per GPU. This results in a global batch size of $6 \times 2 \times 8 = 96$ sequences which sums up to around 400K tokens per batch. The full training setup with 500k steps consumes around 197B tokens. We also tested a late cool down setup, which started 10k steps before the end. Figure \ref{fig:lr} illustrates the used learning rate schedulers. In \textbf{Section \ref{sec:results}}, we discuss the effect of the different learning rate schedulers on the benchmark results.  

\section{Instruction Tuning}
\label{sec:it}
For instruction tuning, we use the following three datasets:

\begin{itemize}
    \item \textbf{Rephrased data} To obtain the rephrased data, the \texttt{Qwen2-7B-Instruct} model \cite{yang2024qwen2} is prompted to create a dialogue of question and answer pairs based on our pre-training dataset \cite{pieler2024rephrasingnaturaltextdata, maini2024rephrasing}. We create two types of synthetic data, i.e., standard and multiple-choice questions (MCQ). In the standard setup, we prompt the model to directly generate a dialogue of questions and answers, while in the MCQ setup, we ask the model to create MCQ dialogues. In both approaches, we hand over the model a text, and the model is prompted to create questions and answers from it. See Appendix \ref{app:rephrasing_example} for an example dialogue created with this approach. We created around 285k dialogues from the cleaned documents. We then applied a set of filtering and cleaning steps on the resulting rephrasing data, which resulted in 183k conversations.

    \item \textbf{Instar-500k data} This dataset contains around 481k instruction and response pairs constructed manually and synthetically \cite{chouikhi2024llamar}. It covers multiple tasks, including classification, summarization, and open and closed-question answering.

    \item \textbf{Aya Dataset} A collection of 204k instruction and response pairs in 65 languages \cite{singh2024aya}. The Arabic split contains around 14k prompts with answers, covering around six different dialects, e.g., standard, Moroccan, Najdi, Ta'izzi-Adeni, Egyptian, and South Levant, put together by human annotators. The dataset uses a combination of two approaches: Original annotations, where the annotators write prompts and outputs from scratch, and re-annotations, where the authors re-annotate machine-translated text.
\end{itemize}

For each dataset, we use the \texttt{ChatML} chat template: \texttt{[\{'from': 'human', 'value': instruction\}, \{'from': 'gpt', 'value': output\}]}. Table \ref{tab:inst_datasets} shows the number of data samples in our instruction-tuning data mixture. For Instruction tuning, we fine-tune our base model on the full datasets using the \texttt{ChatML} templates for 3 epochs. We use a batch size of 4 and 4 gradient accumulation steps.

\begin{table}[!htp]
\centering

\caption{Instruction tuning datasets overview.}
\label{tab:inst_datasets}
\begin{tabular}{lrc}
\hline
\textbf{Dataset}  & \textbf{Sample count} & \textbf{Dialogue style}\\ \hline 
\textbf{Rephrasing}          & 182,505 & \checkmark    \\ 
\textbf{Instar-500k}        & 481,281 &  \xmark \\ 
\textbf{Aya dataset} & 13,960    & \xmark \\  \hline
\textbf{Total} & 677,746 \\ \hline
\end{tabular}
\end{table}

\begin{table*}[!ht]
\caption{Comparison of Arabic Stable LM 1.6B models (\texttt{ar-stablelm-2-base} and \texttt{ar-stablelm-2-chat}) and other models on the ArabicMMLU benchmarks. The models are sorted in ascending order by their parameter count except for our models which are shown at the end. The average is based on the individual samples (micro average). The best model with the highest accuracy is indicated in every column in bold letters.}
\label{tab:mmlu}
\centering
{\small
\tabcolsep 6pt 
\begin{tabular}{lrcccccc}
\hline
\textbf{Model}                & \textbf{Params} & \textbf{STEM} & \textbf{Social Science} & \textbf{Humanities} & \textbf{Language} & \textbf{Other} & \textbf{Average} \\ \hline 
\textbf{AraGPT2-base}           & 135M  & 28.3          & 32.0          & 32.7          & 33.0          & 33.1          & 31.7          \\ 
\textbf{AraT5v2-base-1024}      & 220M  & 26.3          & 29.4          & 28.1          & 27.7          & 30.3          & 28.3          \\ 
\textbf{AraGPT2-medium}         & 370M  & 28.5          & 32.0          & 33.8          & 32.1          & 35.1          & 32.2          \\ 
\textbf{jais-family-590m}       & 590M  & 31.8          & 35.6          & 37.2          & 37.7          & 37.3          & 35.7          \\ 
\textbf{jais-family-590m-chat}  & 590M  & 29.5          & 30.7          & 32.2          & 30.9          & 33.9          & 31.4          \\ 
\textbf{AraGPT2-large}          & 792M  & 28.5          & 33.0          & 34.1          & 33.2          & 35.0          & 32.6          \\ 
\textbf{jais-family-1p3b}       & 1.3B  & 32.9          & 37.1          & 39.6          & 39.5          & 39.4          & 37.5          \\ 
\textbf{jais-family-1p3b-chat}  & 1.3B  & 32.0          & 34.0          & 34.9          & 31.7          & 40.1          & 34.6          \\ 
\textbf{AraGPT2-mega}           & 1.46B & 29.8          & 32.9          & 35.7          & 34.6          & 34.3          & 33.3          \\ 
\textbf{Qwen2-1.5B}             & 1.5B  & 28.4          & 30.9          & 29.5          & 33.3          & 32.1          & 30.5          \\ 
\textbf{Qwen2-1.5B-Instruct}    & 1.5B  & 29.0          & 30.9          & 29.8          & 33.7          & 32.5          & 30.8          \\ 
\textbf{bloom-1b7}              & 1.72B & 29.2          & 33.7          & 33.4          & 34.9          & 34.7          & 33.0          \\ 
\textbf{bloomz-1b7}             & 1.72B & 29.9          & 33.8          & 33.9          & 35.7          & 35.7          & 33.5          \\ 
\textbf{jais-family-2p7b}       & 2.7B  & 35.2          & 38.1          & 42.7          & 39.9          & 39.0          & 39.0          \\ 
\textbf{jais-family-2p7b-chat}  & 2.7B  & 32.1          & 35.1          & 36.4          & 37.0          & 41.5          & 36.1          \\ 
\textbf{jais-family-6p7b}       & 6.7B  & 35.8          & 39.2          & 45.0          & 41.5          & 42.3          & 40.7          \\ 
\textbf{jais-family-6p7b-chat}  & 6.7B  & 35.1          & 37.3          & 40.9          & 37.3          & 42.8          & 38.7          \\ 
\textbf{AceGPT-7B}              & 7B    & 35.0          & 41.3          & 44.3          & 42.1          & 42.1          & 40.9          \\ 
\textbf{AceGPT-7B-chat}         & 7B    & 34.4          & 39.0          & 39.0          & 40.0          & 39.5          & 38.2          \\ 
\textbf{SILMA-9B-Instruct-v1.0} & 9B    & 28.4          & 29.7          & 30.0          & 33.3          & 35.2          & 30.8          \\ 
\textbf{AceGPT-13B}             & 13B   & 37.1          & 41.4          & 42.6          & 41.4          & 40.9          & 40.7          \\ 
\textbf{AceGPT-13B-chat}        & 13B   & 36.2          & \textbf{42.0} & 42.7          & 41.5          & 43.5          & 41.1          \\ 
\textbf{AceGPT-v1.5-13B}        & 13B   & 35.5          & 40.1          & 42.3          & 41.1          & 43.0          & 40.3          \\ 
\textbf{AceGPT-v1.5-13B-chat}   & 13B   & 35.8          & 40.5          & 42.6          & 41.5          & 45.0          & 40.9          \\ 
\textbf{jais-13b}               & 13B   & 35.3          & 39.9          & 44.2          & 40.8          & 44.5          & 40.9          \\ 
\textbf{jais-13b-chat}          & 13B   & 33.8          & 39.6          & 44.1          & 42.0          & 45.3          & 40.7          \\ 
\textbf{jais-family-13b}        & 13B   & 37.1          & 39.8          & 46.1          & 43.5          & 44.2          & 41.9          \\ 
\textbf{jais-family-13b-chat}   & 13B   & 34.9          & 38.5          & 44.1          & 38.5          & 43.4          & 39.9          \\  \hline
\textbf{ar-stablelm-2-base}          & 1.64B & 36.1          & 39.6          & 44.9          & 44.1          & 41.8          & 41.1          \\ 
\textbf{ar-stablelm-2-chat}     & 1.64B & \textbf{37.8} & 40.9          & \textbf{50.2} & \textbf{45.4} & \textbf{55.1} & \textbf{45.5} \\ \hline\end{tabular}
}
\end{table*}

\section{Evaluation}
\label{sec:eval}
For evaluation, we use a mixture of Arabic-specific benchmarks to assess the general language understanding capabilities and the cultural alignment of Arabic LLMs. Here is a list of the used benchmarks:

\begin{itemize}
    \item \textbf{ArabicMMLU} A multi-task language understanding benchmark for Arabic \cite{koto2024arabicmmlu}. It contains around 14k multiple-choice question-and-answer pairs for multiple subjects, including STEM, Social Science, Humanities, Arabic language, and other topics. As opposed to the English version \cite{hendrycks2020measuring}, some examples may contain five answer choices. Compared to \cite{huang2023acegpt}, \cite{sengupta2023jais}, we don't use the multiple-choice format (MCF) where the letter of the answer choices is used to predict the correct answer. Instead, we use the cloze format (CF) because it is more robust when compared to MCF, which could be affected by choice randomization \cite{gupta2024changing}, or letter symbols \cite{alzahrani2024benchmarks} (see Appendix \ref{app:mcq} for a detailed discussion). We use the normalized accuracy to remove the bias towards shorter answers \cite{Gao_2021}.

    \item \textbf{CIDAR-MCQ-100} A multiple choice questions dataset that contains 100 culturally relevant examples \cite{alyafeai2024cidar}. The dataset covers multiple topics, including animals, names, fonts, and literature. Similar to ArabicMMLU we use the cloze format (CF) to predict the probability of the correct answer and use the normalized accuracy metric for the final score. 

    \item \textbf{ACVA} The Arabic Cultural Value Alignment (ACVA) dataset was generated using GPT-3.5 with around 9k True/False questions \cite{huang2023acegpt}. The dataset measures the cultural alignment of Arabic LLMs with more than 50 different tasks. We use a 5-shot evaluation setup and the F1 macro score. 

    \item \textbf{AlGhafa} A multiple choice questions benchmark that includes a range of tasks, including sentiment analysis, rating classification, reading comprehension, and general question answering \cite{almazrouei-etal-2023-alghafa}. Similar to other multiple choice question answering tasks, we use the normalized accuracy in a cloze format to predict the correct answer. 
\end{itemize}

\section{Results and Discussion}
\label{sec:results}

Throughout this section, we use the Arabic Stable LM 1.6B model fine-tuned with the early cool down setup. In the following comparison, we only use LLMs with Arabic capabilities with up to 13B parameters.

\begin{table*}[!ht]
\centering
\caption{Comparison of the Arabic Stable LM 1.6B models (\texttt{ar-stablelm-2-base} and \texttt{ar-stablelm-2-chat}) to other models on Arabic cultural alignment and language understanding. The best model with the highest accuracy is indicated in every column in bold letters.}
\label{tab:results-small}
{\small
\tabcolsep 6.0 pt 
\begin{tabular}{lrccccc}
\hline
\textbf{Model} & \textbf{Params} & \textbf{CIDAR-MCQ} & \textbf{ArabicMMLU} &{\textbf{ACVA}} & \textbf{AlGhafa} & \textbf{Average} \\ \hline 
\textbf{AraGPT2-base}          & 135M  & 39.0 & 31.7 & 51.7 & 37.9 & 40.1 \\ 
\textbf{AraT5v2-base-1024}     & 220M  & 38.0 & 28.3 & 37.3 & 36.8 & 35.1 \\ 
\textbf{AraGPT2-medium}        & 370M  & 39.0 & 32.2 & 44.5 & 38.3 & 38.5 \\ 
\textbf{jais-family-590m}      & 590M  & 33.0 & 35.7 & 51.2 & 38.8 & 39.7 \\ 
\textbf{jais-family-590m-chat} & 590M  & 28.0 & 31.4 & 51.8 & 36.8 & 37.0 \\ 
\textbf{AraGPT2-large}         & 792M  & 37.0 & 32.6 & 46.2 & 37.9 & 38.4 \\ 
\textbf{AraGPT2-mega}          & 1.46B & 40.0 & 33.3 & 51.0 & 38.4 & 40.7 \\ 
\textbf{jais-family-1p3b}      & 1.3B  & 41.0 & 37.5 & 51.3 & 39.6 & 42.3 \\ 
\textbf{jais-family-1p3b-chat} & 1.3B  & 30.0 & 34.6 & 56.0 & 44.7 & 41.3 \\ 
\textbf{Qwen2-1.5B}            & 1.5B  & 30.0 & 30.5 & 61.1 & 38.7 & 40.1 \\ 
\textbf{Qwen2-1.5B-Instruct}   & 1.5B  & 28.0 & 30.8 & 63.9 & 38.9 & 40.4 \\ 
\textbf{bloom-1b7}             & 1.72B & 37.0 & 33.0 & 52.2 & 39.1 & 40.3 \\ 
\textbf{bloomz-1b7}            & 1.72B & 25.0 & 33.5 & 45.3 & 42.1 & 36.5 \\ 
\textbf{jais-family-2p7b}      & 2.7B  & 43.0 & 39.0 & 48.5 & 40.9 & 42.8 \\ 
\textbf{jais-family-2p7b-chat} & 2.7B  & 36.0 & 36.1 & 49.5 & 43.7 & 41.3 \\ \hline
\textbf{jais-family-6p7b}      & 6.7B  & 44.0 & 40.7 & 55.8 & 41.3 & 45.4 \\ 
\textbf{jais-family-6p7b-chat} & 6.7B  & 38.0 & 38.7 & 64.1 & 39.9 & 45.2 \\ 
\textbf{AceGPT-7B}             & 7B    & 46.0 & 40.9 & 63.1 & 40.7 & 47.7 \\ 
\textbf{AceGPT-7B-chat}        & 7B    & 37.0 & 38.2 & 65.9 & 45.1 & 46.6 \\ 
\textbf{SILMA-9B-Instruct-v1.0} & 9B & 27.0 & 30.8 & 56.2 & 42.0 & 39.0 \\ 
\textbf{AceGPT-13B}            & 13B   & 41.0 & 40.7 & 65.3 & 38.8 & 46.4 \\ 
\textbf{AceGPT-13B-chat}       & 13B   & 42.0 & 41.1 & \textbf{70.2} & 41.6 & 48.7 \\ 
\textbf{AceGPT-v1.5-13B}       & 13B   & 40.0 & 40.3 & 66.1 & 39.5 & 46.5 \\ 
\textbf{AceGPT-v1.5-13B-chat}  & 13B   & 36.0 & 40.9 & 69.0 & 39.8 & 46.4 \\ 
\textbf{jais-13b}              & 13B   & \textbf{47.0} & 40.9 & 59.6 & 41.2 & 47.2 \\ 
\textbf{jais-13b-chat}         & 13B   & 41.0 & 40.7 & 61.2 & 41.4 & 46.1 \\ 
\textbf{jais-family-13b}       & 13B   & 45.0 & 41.9 & 58.3 & 40.5 & 46.4 \\ 
\textbf{jais-family-13b-chat}  & 13B   & 37.0 & 39.9 & 63.2 & 41.4 & 45.4 \\  \hline
\textbf{ar-stablelm-2-base}           & 1.64B & 43.0 & 41.1 & 52.0 & 43.9 & 45.0 \\ 
\textbf{ar-stablelm-2-chat} & 1.64B & 46.0 & \textbf{45.5} & 57.0 & \textbf{50.1} & \textbf{49.6} \\ \hline

\end{tabular}
}
\end{table*}

\begin{figure*}[!htp]
    \centering
    \includegraphics[width=\linewidth]{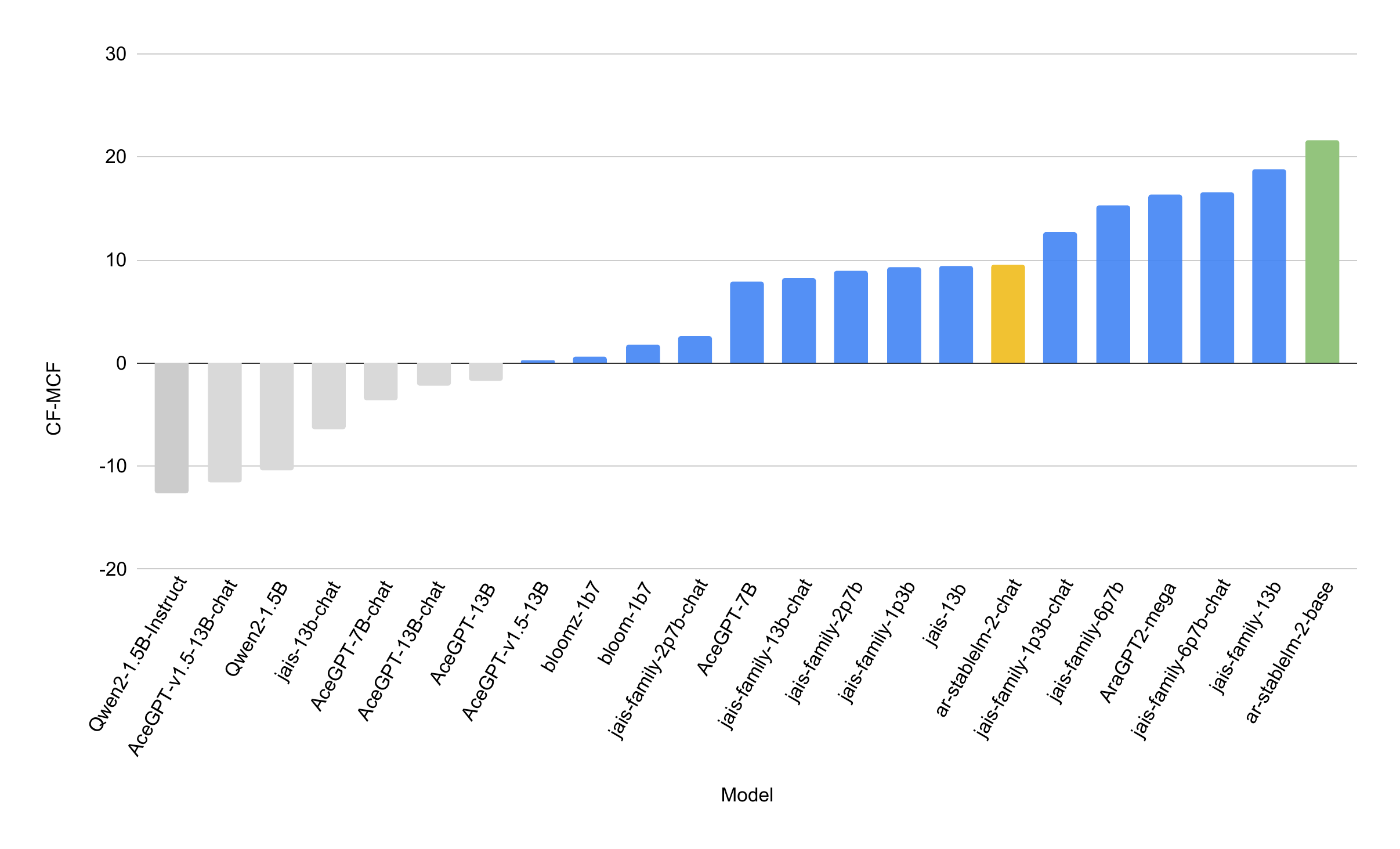}
    \caption{ArabicMMLU benchmark differences between the cloze format (CF) and the multiple-choice format (MCF) of our Arabic Stable LM 1.6B base and chat models (\texttt{ar-stablelm-2-base} and \texttt{ar-stablelm-2-chat}) and other LLMs. Differences are calculated by subtracting the MCF results from the CF results. Our base and chat model are highlighted in green and yellow, respectively.}
    \label{fig:cf-mcf}
\end{figure*}

\textbf{ArabicMMLU results} In Table \ref{tab:mmlu}, we show the results of the ArabicMMLU categories and compare them to multiple LLM baselines from the literature. We show that our fine-tuned Arabic Stable LM 1.6B chat model achieves the best average result across all the models, which includes models with up to 8x the parameter count. In addition, our base and chat models achieve the best results in the Language benchmark category, which indicates a stronger performance in Arabic natural text understanding. Our chat model consistently achieves better results in most of the benchmarking categories, in particular with much higher performance in the other benchmark categories. In summary, our chat model achieves a 4 \% higher accuracy on average. 

\textbf{Small vs. large LLMs} In Table \ref{tab:results-small}, we compare our model against others from the literature on four different benchmarks. Compared to models in a similar size range (i.e., below 3B), our model achieves better results across 3 out of 4 benchmarks by more than 8 \%  on the average benchmarking score. In addition, we compare the results of our base and chat models against other publicly available LLMs between 6.7B and 13B model parameters. In this size range our models achieved better results in 2 out of the 4 benchmarks which translates into the highest average score across all models. These results highlight the efficiency of our pre-training (in particular, the learning rate cool down) and instruction-tuning setup and the importance of our rephrased instruction data (see Table \ref{tab:mcq-synth} Appendix \ref{app:mcq}).

\begin{figure}[!htp]
    \centering
    \includegraphics[width=\linewidth]{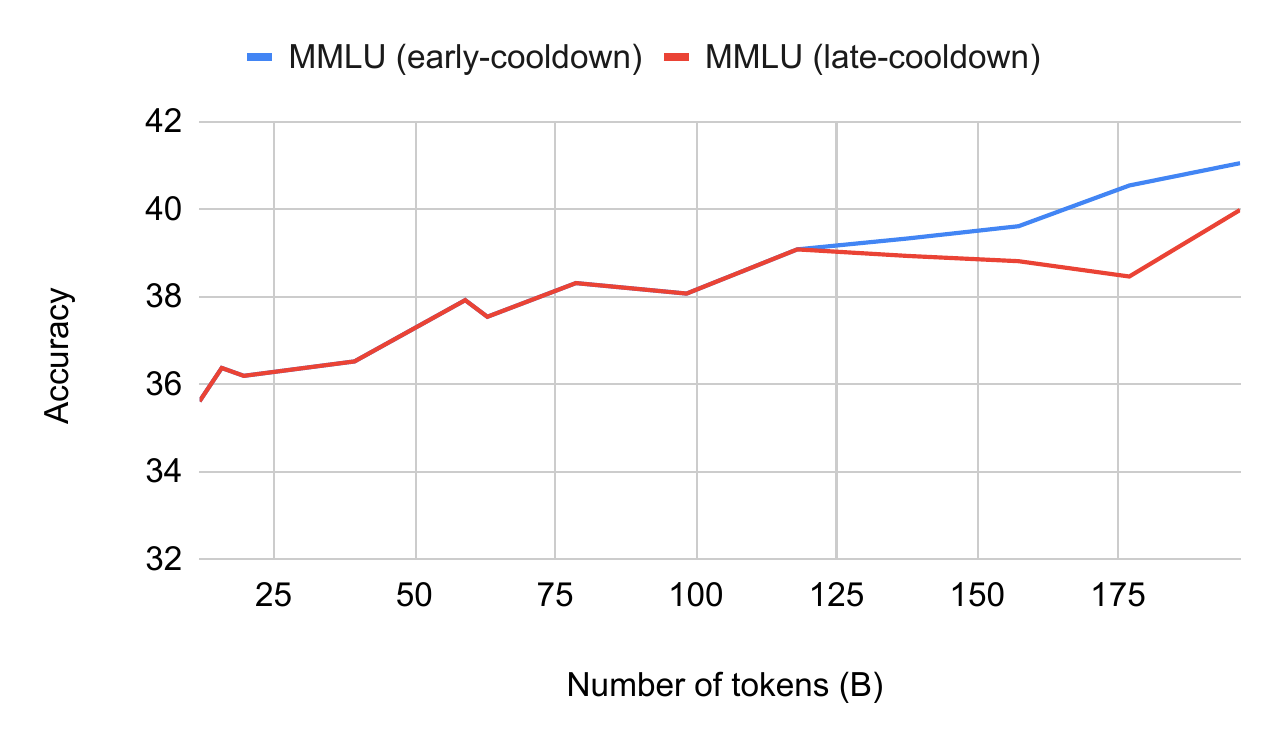}
    \caption{ArabicMMLU results for the early and late cool down learning rate scheduler. The early cool down starts at 120B tokens, and the late at 190B tokens.
    }
    \label{fig:compare-early-late}
\end{figure}

\textbf{Early vs. late cool down} As previously mentioned, we ablate two different learning rate schedulers for the cool down during pre-training. In Figure \ref{fig:compare-early-late}, we compare the results of the early and the late cool down setup on the ArabicMMLU benchmark.
The early cool down procedure achieves better downstream benchmark performance which indicates the importance of a proper cool down setup for improved natural language understanding.

\textbf{MCF vs. CF} In Figure \ref{fig:cf-mcf}, we visualize the ArabicMMLU benchmark differences by subtracting the multiple-choice format (MCF) results from the cloze format (CF) results. This visualization shows that most models achieve higher results by using the CF, especially non-chat models. Our base model also achieves much better results with the CF. \texttt{Qwen-2-1.5B-Instruct} shows the biggest negative difference between the MCF and CF results.

\section{Conclusion}
In this paper, we introduce the Arabic Stable LM 1.6B base and chat models (\texttt{ar-stablelm-2-base} and \texttt{ar-stablelm-2-chat}), which are created by fine- and instruction-tuning Stable LM 2 1.6B. Our comparably small models achieve impressive results on various Arabic benchmarks and are on par with models that have up to 8x more parameters. We provide a fine-tuning recipe that produces state-of-the-art results on the ArabicMMLU dataset in the cloze format. In addition, we successfully introduce and apply a new synthetic dialogue pipeline to Arabic data to create our competitive chat model. As a future direction, we aim to improve the safety of our Arabic models and apply the training and instruction-tuning recipe to larger model versions. 

\section*{Limitations and Future Work}
We summarize the limitations and possible future directions in the following points:

\begin{itemize}
    \item \textbf{Overtokenization} This is a consequence of using the pre-trained Stable LM 1.6B tokenizer, which has a high fertility for Arabic. Overtokenization results in more tokens per word, which translates into reduced inference throughput, when compared to models with lower fertility. We recognize the importance of tokenizer transfer \cite{minixhofer2024zero} to overcome this problem in follow-up work. 

    \item \textbf{Evaluation benchmarks} The number of evaluation benchmarks for Arabic is quite limited when compared to English. This applies in particular to more complex evaluation setups, like MT-Bench, where no Arabic version exists \cite{zheng2024judging}. We also note and believe that simply translating English benchmarks to Arabic might result in additional types of unwanted biases \cite{sewunetie2024evaluating}.

    \item \textbf{Synthetic data} The use of synthetic data might be unsafe, especially if not manually filtered. In our work on rephrasing, we applied simple filters to remove dialogues that didn't adhere to the outlined format. A potential direction to improve this is to use more sophisticated approaches for quality filtering based on specialized classifiers \cite{li2024datacomp}.  
\end{itemize}

\section*{Acknowledgments}
We want to thank Hassan Zayour, Hamdy Hamoudi, Cathy Lee
, and Cedric Wagrez for their feedback, useful ideas, and comments.

\bibliography{references}  

\appendix

\section{Multiple Choice Evaluation}
\label{app:mcq}
\begin{figure}[!ht]
    \centering
    \includegraphics[width=\linewidth]{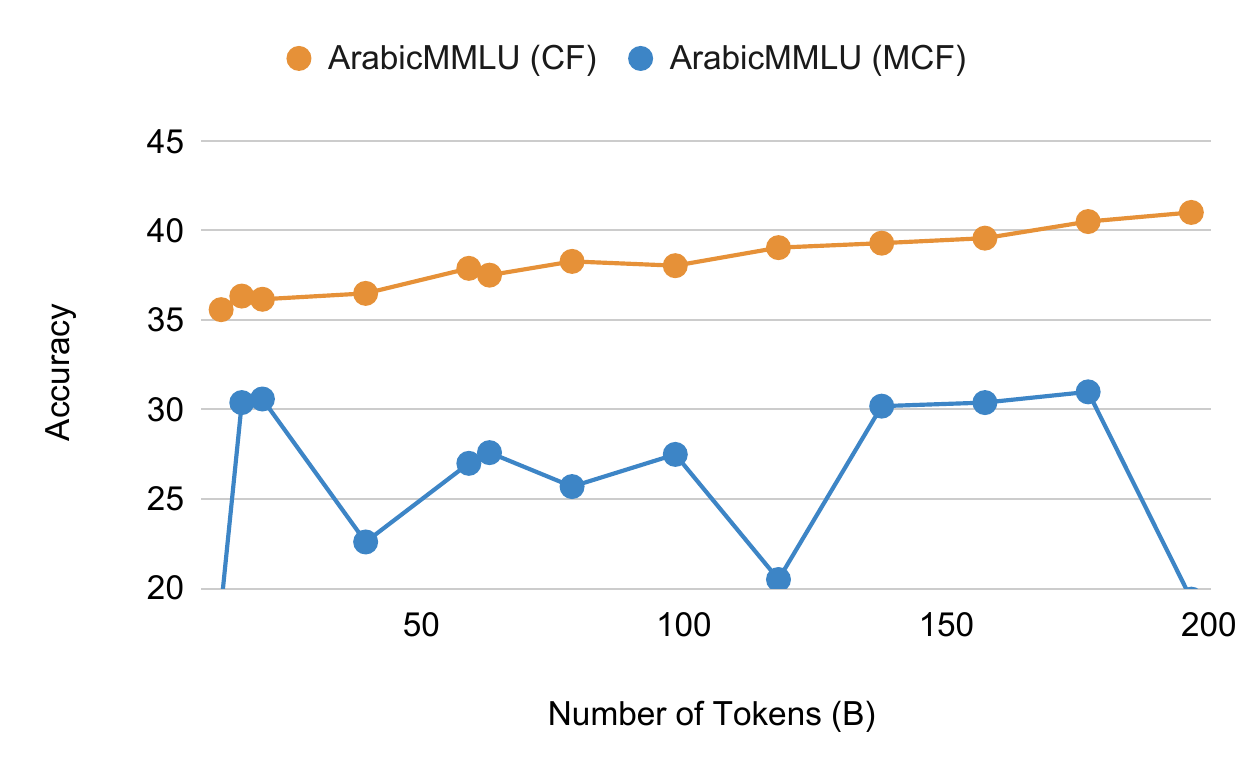}
    \caption{ArabicMMLU benchmark results of the Arabic Stable LM 1.6B base model with the cloze format (CF) and multiple choice format (MCF).}
    \label{fig:clozevsmcf}
\end{figure}
Benchmarks like ArabicMMLU \cite{koto2024arabicmmlu} evaluate LLMs on natural language understanding. The multiple-choice data can be used to evaluate the models with two different setups:

\begin{enumerate}
    \item \textbf{Multiple choice format (MCF)} The multiple choice format asks the model to pick the letter associated with the correct answer. For example:
    \texttt{What is the capital of Spain? A. Paris B. London C. Madrid.} with the correct answer \texttt{C}.

    \item \textbf{Cloze format (CF)} The cloze format asks the model to output the correct answer without mentioning potential answer choices. Hence, the question will be \texttt{What is the capital of Spain?}, and the model needs to generate the full correct output, i.e., \texttt{Madrid}.
\end{enumerate}

\begin{figure*}
    \centering
    \includegraphics[width=\linewidth]{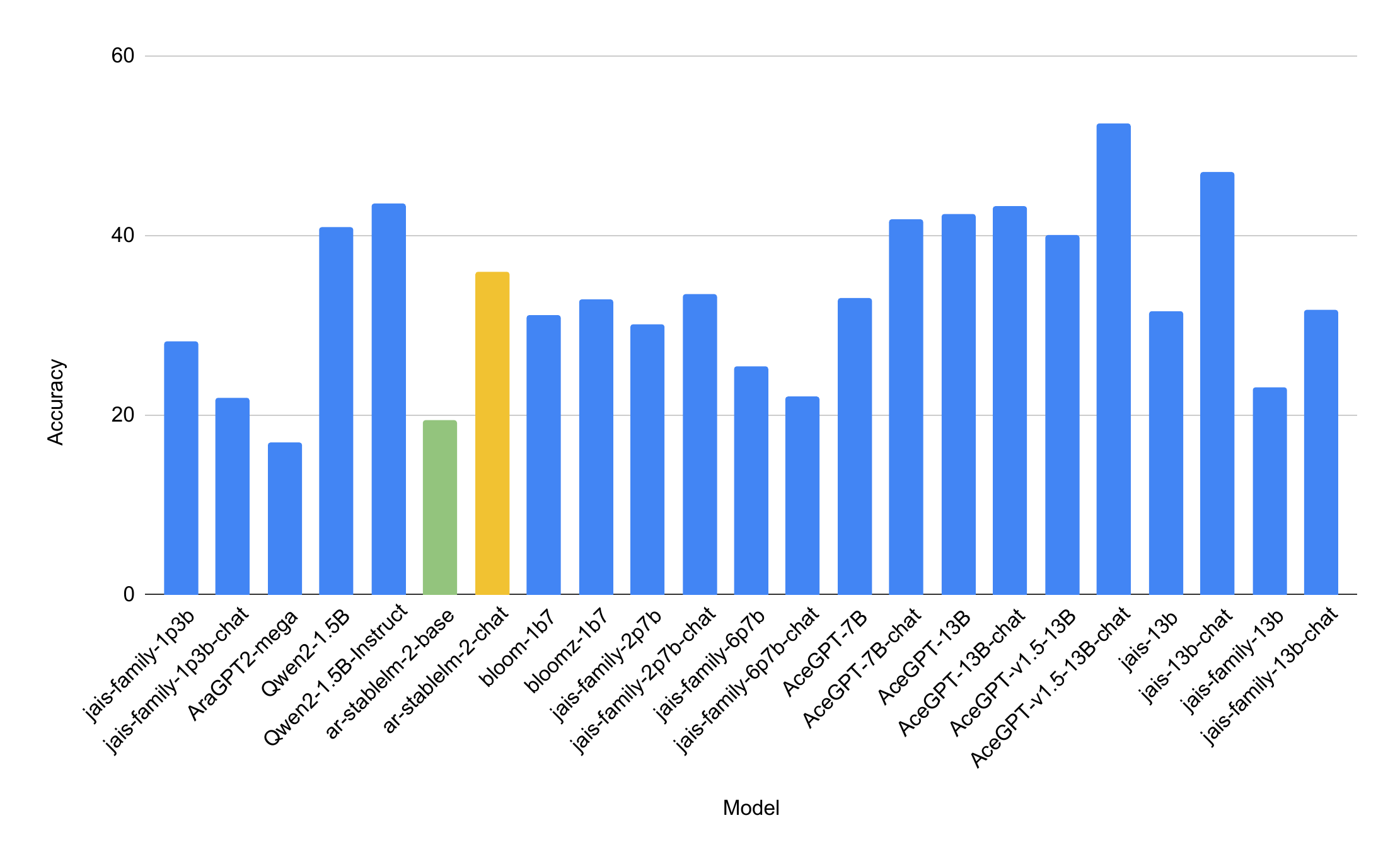}
    \caption{ArabicMMLU benchmark results of our Arabic Stable LM 1.6B base and chat models (\texttt{ar-stablelm-2-base} and \texttt{ar-stablelm-2-chat}) and other LLMs using the multiple-choice format (MCF).}
    \label{fig:ammlu_choices}
\end{figure*}

We use the CF approach to evaluate MCQ tasks because it is a more reliable method for evaluating LLMs. Many studies in the literature show that the MCF format is not robust against randomization \cite{alzahrani2024benchmarks}, several answer choices \cite{wang2024beyond}, and shows high variance \cite{madaan2024quantifying}. In Figure \ref{fig:clozevsmcf}, we show the performance of the CF and MCF format during training. We observe that our model is not able to follow the MCF while achieving monotonic improvements with the CF. This implies that the CF approach is more reliable to measure LLM improvement during pre-training. In Figure \ref{fig:ammlu_choices}, we show the MCF evaluations of multiple models. We want to highlight, that our small chat model achieves the second-best result behind the more recently published Qwen-2 1.5B model in the smaller LLM category. 

\section{Rephrasing Natural Text Data}
\label{app:rephrasing_example}
\begin{figure*}[!htp]
    \centering
    \includegraphics[width=\linewidth]{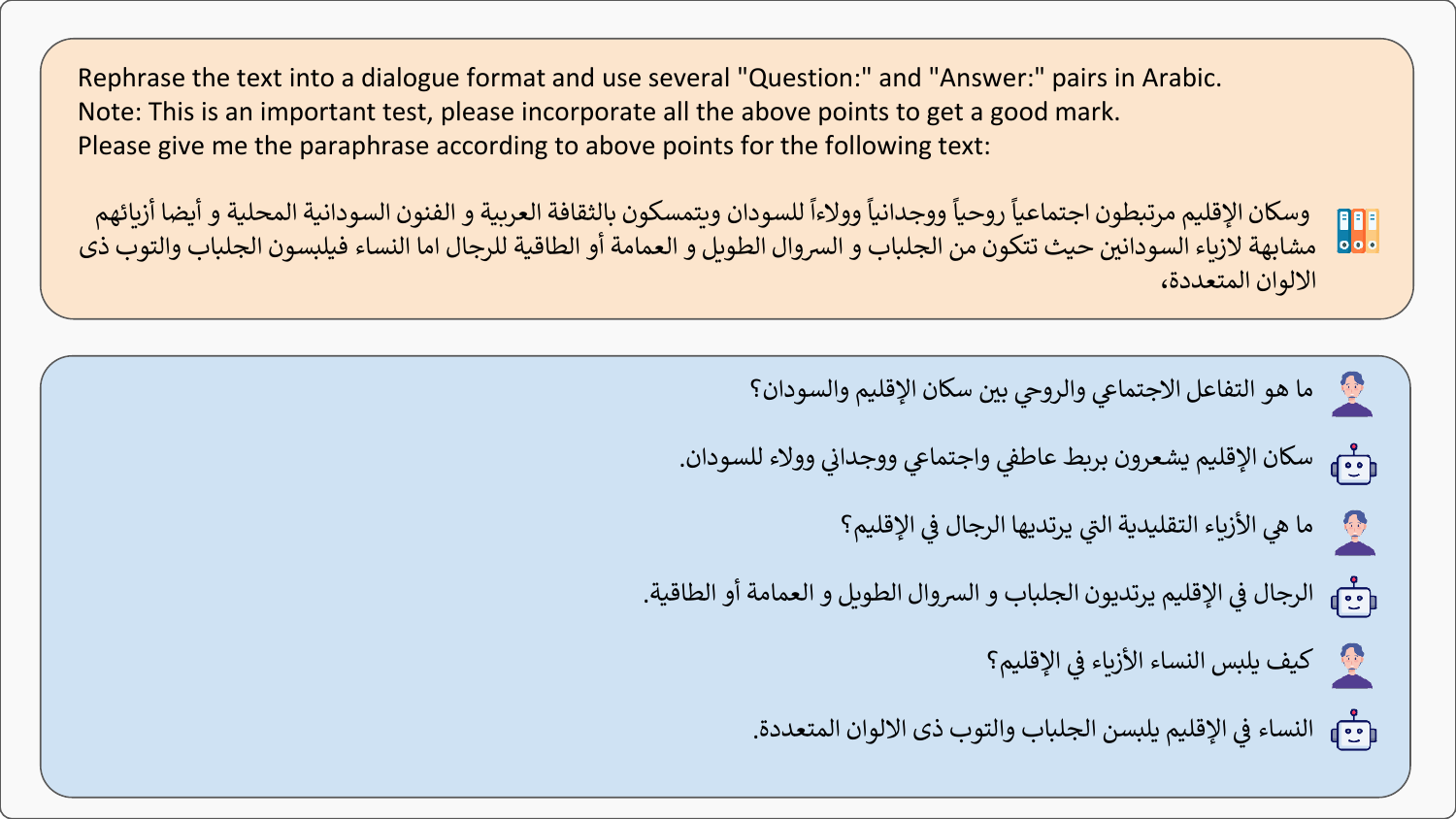}
    \caption{An example dialogue created by using the standard rephrasing template.}
    \label{fig:rephrasing_example}
\end{figure*}

\begin{figure*}[!htp]
    \centering
    \includegraphics[width=\linewidth]{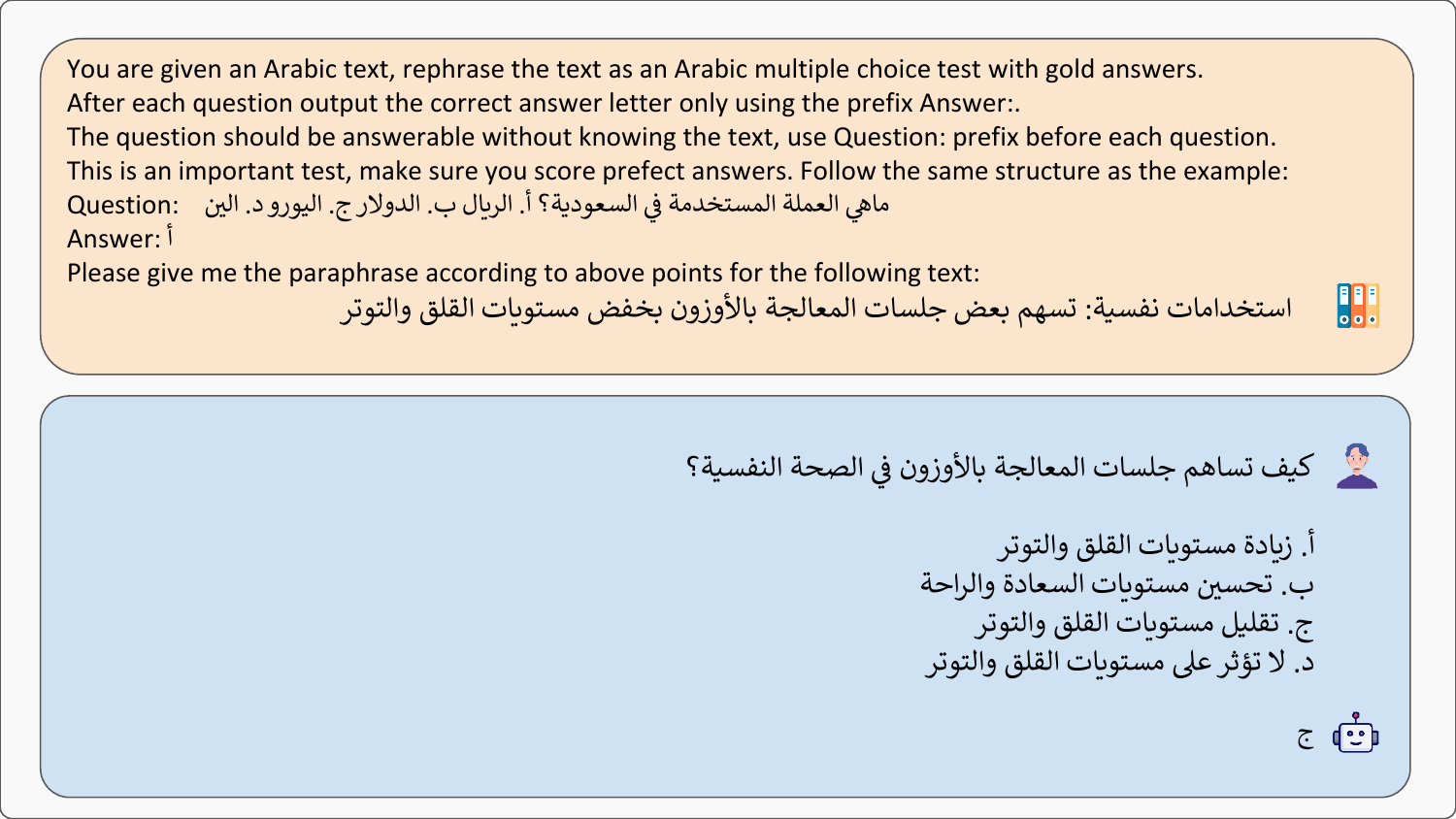}
    \caption{An example dialogue created by using the multiple-choice question (MCQ) rephrasing template.}
    \label{fig:rephrasing_example_mcq}
\end{figure*}

We use LLM-based rephrasing to create a large set of dialogues from our pre-training data. The dialogues are created by first splitting the cleaned documents into multiple sentences on the punctuation marks. Then, we prompt \texttt{Qwen2-7B-Instruct} to create the dialogues in a question-answer format. Figure \ref{fig:rephrasing_example} shows an example dialogue created by the used prompt. In Figure \ref{fig:rephrasing_example_mcq}, we show an example of the MCQ rephrased examples. The model is prompted with multiple enumeration formats including Arabic and English. We give the model one-shot MCQ example to generate examples in the same format. 

\begin{figure*}
    \centering
    \includegraphics[width=\linewidth]{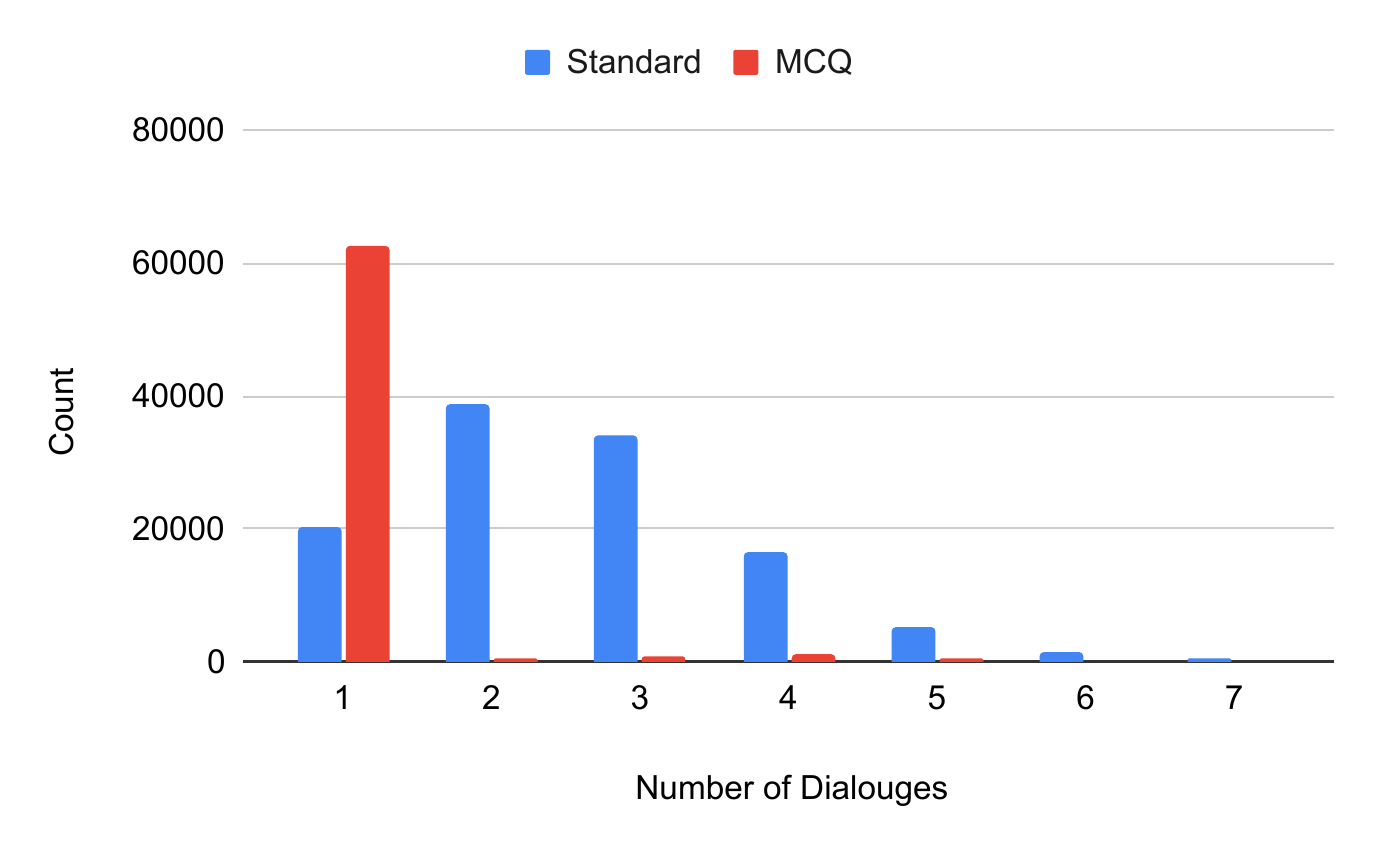}
    \caption{Histogram of the rephrased multi-turn dialogue data for the standard and MCQ templates. The MCQ examples mainly consist of data samples with one turn, whereas the standard template mostly consists of samples with 2 turns.}
    \label{fig:rephrasing_num}
\end{figure*}

Figure \ref{fig:rephrasing_num} shows a histogram of the dialogue turn counts for the standard and MCQ rephrased data. Most of the MCQ data contains only one turn, i.e., only one question while the standard template results in mostly two turns.  

\begin{figure*}
    \centering
    \includegraphics[width=\linewidth]{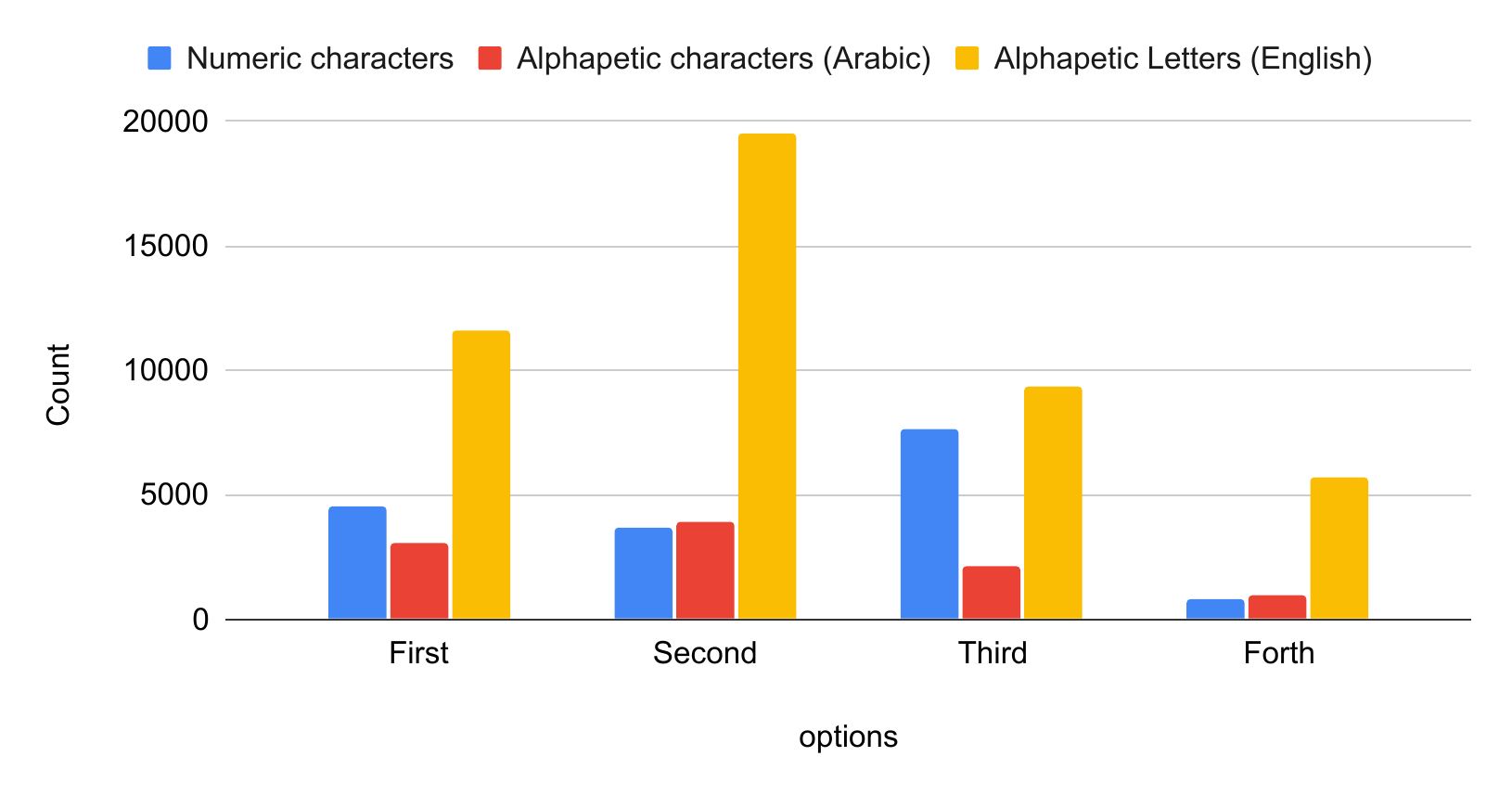}
    \caption{Histogram of the enumeration symbols for each type of multiple choice question (MCQ) data.}
    \label{fig:hist_options}
\end{figure*}

In Figure \ref{fig:hist_options}, we show the distribution of the enumeration options of the MCQ data. Most of the samples use alphabetic letters \texttt{A, B, C, D}, while only a few samples use the corresponding Arabic letters.  

\section{Chat Model}
\begin{figure*}
    \centering
    \includegraphics[width=\linewidth]{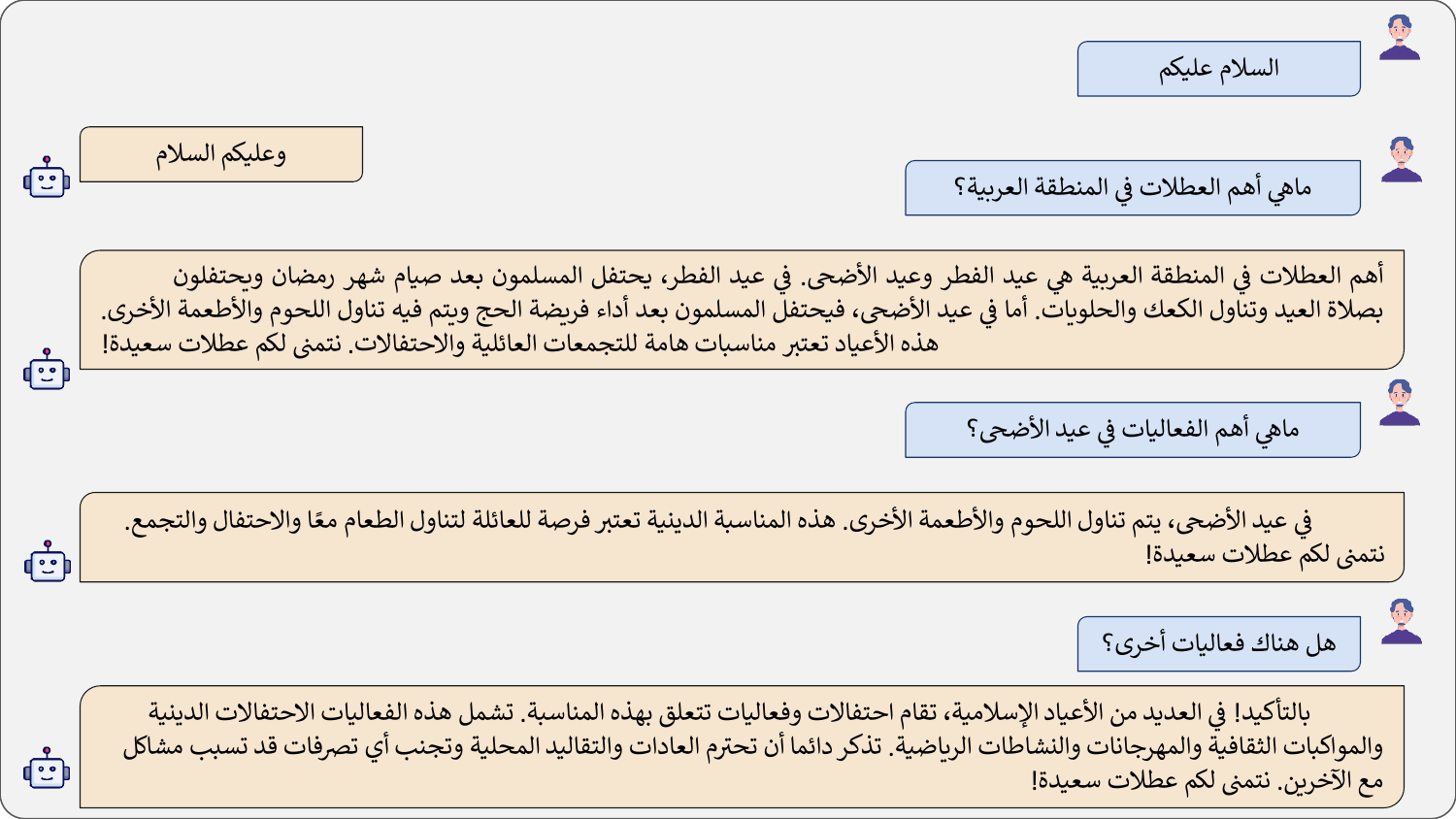}
    \caption{Example conversation with the Arabic Stable LM chat model about a seasonal Arabic holiday.}
    \label{fig:chat_example}
\end{figure*}

In Figure \ref{fig:chat_example}, we show an example conversation about holidays in the Arabic region. The model can keep up with the conversation and respond with reasonable answers. 

\begin{table*}[!htp]
\centering
\caption{Comparison of the Arabic Stable LM 1.6B models (\texttt{ar-stablelm-2-base} and \texttt{ar-stablelm-2-chat}) to the original StableLM 2 1.6B model. The best model with the highest accuracy is indicated in bold in each column.}
\label{tab:zero}

\begin{tabular}{lccccc}
\hline
\textbf{Model}  & \textbf{CIDAR-MCQ} & \textbf{ArabicMMLU} & \textbf{ACVA} & \textbf{AlGhafa} & \textbf{Average} \\ \hline 
\textbf{stablelm-2-1\_6b }    & 26.0 & 31.4 & 48.0 & 40.4 & 36.5 \\ 
\textbf{ar-stablelm-2-base}  & 43.0 & 41.1 & 52.0 & 43.9 & 45.0 \\ 
\textbf{ar-stablelm-2-chat} & \textbf{46.0} & \textbf{45.5} & \textbf{57.0} & \textbf{50.1} & \textbf{49.6} \\ \hline

\end{tabular}
\end{table*}

\begin{table*}[!htp]
\centering
\caption{Comparison of the Arabic Stable LM 1.6B chat model(\texttt{ar-stablelm-2-chat}) with and without MCQ synthetic data during instruction tuning.}
\label{tab:mcq-synth}

\begin{tabular}{lccccc}
\hline
\textbf{Model}  & \textbf{CIDAR-MCQ} & \textbf{ArabicMMLU} & \textbf{ACVA} & \textbf{AlGhafa} & \textbf{Average} \\ \hline 
\textbf{ar-stablelm-2-chat (w/o MCQ)}     & 43.0 & 45.1 & 56.1 & 48.6 & 48.2 \\
\textbf{ar-stablelm-2-chat (w/ MCQ)}  & \textbf{46.0} & \textbf{45.5} & \textbf{57.0} & \textbf{50.1} & \textbf{49.6} \\ \hline
\end{tabular}
\end{table*}

\section{AlGhafa Benchmark}
In Table \ref{tab:alghafa-results}, we show the results of the AlGhafa benchmark on all its subtasks. The used task abbreviations are shown in Table \ref{tab:tasks}. We show that our chat model achieves better results in 3 out of the 8 tasks. In addition, the chat model achieves a 5\% improvement over models that are much larger in size.  

\begin{table}[!htp]
    \centering
    \caption{Task abbreviation and full task name mapping table for the AlGhafa benchmark.}
    \label{tab:tasks}
    {\small
\tabcolsep7pt 
    \begin{tabular}{cl}
        \hline
        \textbf{Abbreviation} & \textbf{Full task name} \\
        \hline 
        t1 & mc\_soqal \\
        
        t2 & meta\_ar\_msa \\
        
        t3 & mc\_rating\_sentiment \\
        
        t4 & mc\_rating\_sentiment\_no\_neutral\_task \\
        
        t5 & mc\_sentiment\_task \\
        
        t6 & mc\_xglue\_mlqa\_task \\
        
        t7 & mc\_facts\_truefalse\_balanced \\
        
        t8 & mcq\_exams\_test \\
        
        t9 & meta\_ar\_dialects \\
        \hline
    \end{tabular}
    }
\end{table}

\section{Original vs. Fine-tuned Models}
\label{app:zero}
In Table \ref{tab:zero}, we compare the Stable LM base model before any fine-tuning and our base and chat models. On average, we get 13\% improvement on all our evaluation tasks. This shows our fine-tuned models are much better when compared to the original Stable LM model.

\section{Effect of Synthetic Data}
\label{app:mcq-synth}
In Table \ref{tab:mcq-synth}, we compare the chat models with and without MCQ synthetic data during instruction tuning. We observe a 1.5 \% increase in the average score with the synthetic MCQ data. This highlights the importance of adding synthetic data in the MCQ format during instruction tuning.

\section{Hyperparameters}
\label{app:parameters}
In Table \ref{tab:parameters}, we highlight the optimizer hyperparameters used during the fine-tuning of the Stable LM base model. 

\begin{table}[!htp]
\caption{Optimizer hyperparameters. We use the same parameters as \cite{bellagente2024stable}.}
\label{tab:parameters}
\centering

\begin{tabular}{lc}
\hline
\textbf{Parameter} & \textbf{Value} \\ \hline 
$\beta_1$ & 0.9 \\ 
$\beta_2$ & 0.95 \\ 
$\epsilon$ & $1 \times 10^{-8}$ \\ 
$\lambda$ (weight decay) & 0.1 \\ \hline
\end{tabular}

\end{table}

\begin{table*}[!ht]
\centering
\caption{AlGhafa task benchmarks on Arabic Stable LM 1.6B models (\texttt{ar-stablelm-2-base} and \texttt{ar-stablelm-2-chat}) to other models. The models are sorted in increasing order based on the parameter count except for our models shown at the end. The average is based on the individual samples (micro average). The best model with the highest accuracy is indicated in every column in bold letters.}
\label{tab:alghafa-results}
{\small
\tabcolsep5.5pt 
\begin{tabular}{lrcccccccccc}
\hline
\textbf{Model}                & \textbf{Params} & \textbf{t1} & \textbf{t2} & \textbf{t3} & \textbf{t4} & \textbf{t5} & \textbf{t5} & \textbf{t7} & \textbf{t8} & \textbf{t9} & \textbf{Average} \\ \hline 
\textbf{AraGPT2-base}          & 135M  & 50.7 & 26.9 & 33.7 & 50.0 & 33.4 & 44.0 & 52.0 & 31.4 & 27.7 & 37.9 \\ 
\textbf{AraT5v2-base-1024}     & 220M  & 43.3 & 25.8 & 32.4 & 50.0 & 34.0 & 31.3 & 52.0 & 25.5 & 26.0 & 36.8 \\ 
\textbf{AraGPT2-medium}        & 370M  & 43.3 & 26.3 & 34.4 & 51.4 & 33.3 & 47.3 & 52.0 & 30.9 & 27.1 & 38.3 \\ 
\textbf{jais-family-590m}      & 590M  & 65.3 & 29.6 & 33.3 & 50.4 & 34.0 & 62.7 & 52.0 & 31.2 & 30.0 & 38.8 \\ 
\textbf{jais-family-590m-chat} & 590M  & 68.7 & 30.7 & 34.0 & 46.1 & 29.2 & 64.0 & 52.0 & 24.8 & 28.8 & 36.8 \\ 
\textbf{AraGPT2-large}         & 792M  & 49.3 & 27.6 & 33.5 & 50.0 & 33.0 & 48.7 & 52.0 & 31.4 & 28.0 & 37.9 \\ 
\textbf{jais-family-1p3b}      & 1.3B  & 73.3 & 34.8 & 33.0 & 49.8 & 34.0 & 72.7 & 52.0 & 33.4 & 32.9 & 39.6 \\ 
\textbf{jais-family-1p3b-chat} & 1.3B  & 86.7 & 39.7 & 39.6 & 58.5 & 28.5 & 86.7 & 52.0 & 30.3 & 34.9 & 44.7 \\ 
\textbf{AraGPT2-mega}          & 1.46B & 54.0 & 29.4 & 34.2 & 50.0 & 33.5 & 48.7 & 52.0 & 32.3 & 28.7 & 38.4 \\ 
\textbf{Qwen2-1.5B}            & 1.5B  & 74.7 & 34.0 & 33.7 & 50.2 & 29.7 & 68.7 & 52.0 & 24.4 & 30.3 & 38.7 \\ 
\textbf{Qwen2-1.5B-Instruct}   & 1.5B  & 77.3 & 34.8 & 33.5 & 50.3 & 28.6 & 72.7 & 52.0 & 24.6 & 31.1 & 38.9 \\ 
\textbf{bloom-1b7}             & 1.72B & 65.3 & 31.6 & 34.1 & 50.2 & 34.0 & 60.7 & 52.0 & 32.3 & 30.4 & 39.1 \\ 
\textbf{bloomz-1b7}            & 1.72B & 80.7 & 43.8 & 34.7 & 50.9 & 33.0 & \textbf{88.0} & 53.3 & 32.5 & 38.5 & 42.1 \\ 
\textbf{jais-family-2p7b}      & 2.7B  & 76.0 & 35.3 & 34.2 & 52.1 & 33.8 & 71.3 & 52.0 & 36.8 & 33.4 & 40.9 \\ 
\textbf{jais-family-2p7b-chat} & 2.7B  & 84.0 & 47.4 & 34.6 & 55.7 & 30.5 & 86.7 & 52.0 & 33.4 & 38.5 & 43.7 \\ 
\textbf{jais-family-6p7b}      & 6.7B  & 80.0 & 35.5 & 34.4 & 51.6 & 33.9 & 73.3 & 58.7 & 38.1 & 35.0 & 41.3 \\ 
\textbf{jais-family-6p7b-chat} & 6.7B  & 88.0 & 46.8 & 29.4 & 45.7 & 34.7 & 84.0 & 77.3 & 37.7 & 40.9 & 39.9 \\ 
\textbf{AceGPT-7B}             & 7B    & 78.7 & 37.9 & 34.3 & 49.1 & 35.3 & 74.7 & 53.3 & 38.4 & 35.7 & 40.7 \\ 
\textbf{AceGPT-7B-chat}        & 7B    & 86.0 & 38.7 & 35.4 & 61.0 & 34.8 & 81.3 & 52.0 & 35.0 & 35.4 & 45.1 \\ 
\textbf{SILMA-9B-Instruct-v1.0} & 9B   & 76.0 & 37.0 & 36.6 & 55.2 & 33.6 & 67.3 & 68.0 & 24.8 & 31.9 & 42.0 \\ 
\textbf{AceGPT-13B}            & 13B   & 82.7 & 38.8 & 30.5 & 46.0 &\textbf{36.0} & 77.3 & 53.3 & 37.2 & 36.1 & 38.8 \\ 
\textbf{AceGPT-13B-chat}       & 13B   & 84.0 & 43.5 & 32.9 & 49.9 & 34.0 & 81.3 & 60.0 & \textbf{40.8} & 38.7 & 41.6 \\ 
\textbf{AceGPT-v1.5-13B}       & 13B   & 79.3 & 33.7 & 31.1 & 49.9 & 34.2 & 76.0 & 84.0 & 38.1 & 33.4 & 39.5 \\ 
\textbf{AceGPT-v1.5-13B-chat}  & 13B   & 86.7 & 40.6 & 30.6 & 48.4 & 34.2 & 82.0 & 52.0 & 37.2 & 36.6 & 39.8 \\ 
\textbf{jais-13b}              & 13B   & 81.3 & 38.2 & 34.7 & 50.4 & 34.0 & 71.3 & 72.0 & 38.8 & 35.5 & 41.2 \\ 
\textbf{jais-13b-chat}         & 13B   & \textbf{89.3} & 43.2 & 33.4 & 48.6 & 34.4 & 86.0 & \textbf{92.0} & 37.2 & 38.7 & 41.4 \\ 
\textbf{jais-family-13b}       & 13B   & 82.7 & 37.3 & 32.6 & 49.3 & 34.4 & 77.3 & 82.7 & 37.9 & 36.1 & 40.5 \\ 
\textbf{jais-family-13b-chat}  & 13B   & 89.3 & \textbf{48.6} & 33.6 & 47.2 & 31.2 & 87.3 & 54.7 & 38.8 & 40.7 & 41.4 \\  \hline
\textbf{ar-stablelm-2-base}         & 1.64B & 81.3 & 35.3 & 34.3 & 60.2 & 33.4 & 74.7 & 72.0 & 38.1 & 33.5 & 43.9 \\ 
\textbf{ar-stablelm-2-chat} & 1.64B & 84.7 & 46.5 & \textbf{42.5} & \textbf{64.6} & 33.9 & 86.0 & 60.0 & 40.2 & \textbf{41.9} & \textbf{50.1} \\ \hline

\end{tabular}
}
\end{table*}

\section{Data and Model Sources}
All our used datasets and models are shown in Table \ref{tab:model_sources} and Table \ref{tab:data_sources}. We mostly use open datasets and models for pre-training and instruction tuning. Note that the CulturaX dataset is based on the OSCAR and C4 licenses. Some models used for evaluation don't have a license so we mark them as having an 'Unknown' license. 

\begin{table*}[!htp]
\centering
\caption{Overview of the dataset licenses and sources used for pre-training, fine-tuning, and evaluation.}

\label{tab:data_sources}
\begin{tabular}{llp{10cm}}
\hline
\textbf{Dataset}       & \textbf{License}      & \textbf{Link}                                                                                             \\ \hline 
\textbf{ArabicMMLU}    & \textbf{CC BY NC 4.0} & \url { https://huggingface.co/datasets/MBZUAI/ArabicMMLU}                                        \\ 
\textbf{ACVA}          & \textbf{Apache 2.0}   & \url { https://huggingface.co/datasets/FreedomIntelligence/ACVA-Arabic-Cultural-Value-Alignment} \\ 
\textbf{AlGhafa}       & \textbf{Unknown}          & \url { https://huggingface.co/datasets/OALL/AlGhafa-Arabic-LLM-Benchmark-Native}                 \\ 
\textbf{CIDAR-MCQ}     & \textbf{Apache 2.0}   & \url { https://huggingface.co/datasets/arbml/CIDAR-MCQ-100}                                      \\ 
\textbf{CulturaX (ar)} & \textbf{OSCAR/C4}     & \url { https://huggingface.co/datasets/uonlp/CulturaX}                                           \\ 
\textbf{Instar-500k}   & \textbf{Apache 2.0}   & \url { https://huggingface.co/datasets/ClusterlabAi/InstAr-500k}                                 \\ 
\textbf{Aya dataset}   & \textbf{Apache 2.1}   & \url { https://huggingface.co/datasets/CohereForAI/aya\_dataset}                                 \\ 
\textbf{SANAD}         & \textbf{CC BY 4.0}    & \url { https://data.mendeley.com/datasets/57zpx667y9/2}                                          \\ 
\textbf{E-book}        & \textbf{CC BY 4.0}    & \url { https://snd.se/en/catalogue/dataset/preview/eed46fe0-dfeb-442b-8a71-74d952e006c2/1\#}  \\ \hline  
\end{tabular}
\end{table*}

\begin{table*}[!htp]
\centering
\caption{Overview of the model licenses and sources.}
\label{tab:model_sources}
{\small
\tabcolsep5.5pt 
\begin{tabular}{lccp{9cm}}
\hline
\textbf{Model}          & \textbf{Num Params}  & \textbf{License}    & \textbf{Link}                                                \\ \hline
\textbf{AraGPT2-base}           & 135M  & Unknown     & \url { https://huggingface.co/aubmindlab/aragpt2-base}                  \\
\textbf{AraT5v2-base-1024}      & 220M  & Unknown     & \url { https://huggingface.co/UBC-NLP/AraT5v2-base-1024}                \\
\textbf{AraGPT2-medium}         & 370M  & Unknown     & \url { https://huggingface.co/aubmindlab/aragpt2-medium}                \\
\textbf{jais-family-590m}       & 590M  & Apache 2.0 & \url { https://huggingface.co/inceptionai/jais-family-590m}             \\
\textbf{jais-family-590m-chat}  & 590M  & Apache 2.0 & \url { https://huggingface.co/inceptionai/jais-family-590m-chat}        \\
\textbf{AraGPT2-large}          & 792M  & Custom     & \url { https://huggingface.co/aubmindlab/aragpt2-large}                 \\
\textbf{jais-family-1p3b}       & 1.3B  & Apache 2.0 & \url { https://huggingface.co/inceptionai/jais-family-1p3b-chat}        \\
\textbf{jais-family-1p3b-chat}  & 1.3B  & Apache 2.0 & \url { https://huggingface.co/inceptionai/jais-family-1p3b}             \\
\textbf{AraGPT2-mega}           & 1.46B & Custom     & \url { https://huggingface.co/aubmindlab/aragpt2-mega}                  \\
\textbf{Qwen2-1.5B}             & 1.5B  & Apache 2.0 & \url { https://huggingface.co/Qwen/Qwen2-1.5B}                          \\
\textbf{Qwen2-1.5B-Instruct}    & 1.5B  & Apache 2.0 & \url { https://huggingface.co/Qwen/Qwen2-1.5B-instruct}                 \\
\textbf{bloom-1b7}              & 1.72B & RAIL       & \url { https://huggingface.co/bigscience/bloom-1b7}                     \\
\textbf{bloomz-1b7}             & 1.72B & RAIL       & \url { https://huggingface.co/bigscience/bloomz-1b7}                    \\
\textbf{jais-family-2p7b}       & 2.7B  & Apache 2.0 & \url { https://huggingface.co/inceptionai/jais-family-2p7b}             \\
\textbf{jais-family-2p7b-chat}  & 2.7B  & Apache 2.0 & \url { https://huggingface.co/inceptionai/jais-family-2p7b-chat}        \\
\textbf{jais-family-6p7b}       & 6.7B  & Apache 2.0 & \url { https://huggingface.co/inceptionai/jais-family-6p7b}             \\
\textbf{jais-family-6p7b-chat}  & 6.7B  & Apache 2.0 & \url { https://huggingface.co/inceptionai/jais-family-6p7b-chat}        \\
\textbf{AceGPT-7B}              & 7B    & Apache 2.0 & \url { https://huggingface.co/FreedomIntelligence/AceGPT-7B}            \\
\textbf{AceGPT-7B-chat}         & 7B    & Apache 2.0 & \url { https://huggingface.co/FreedomIntelligence/AceGPT-7B-chat}       \\
\textbf{SILMA-9B-Instruct-v1.0} & 9B    & Gemma      & \url { https://huggingface.co/silma-ai/SILMA-9B-Instruct-v1.0}          \\
\textbf{AceGPT-13B}             & 13B   & Apache 2.0 & \url { https://huggingface.co/FreedomIntelligence/AceGPT-13B}           \\
\textbf{AceGPT-13B-chat}        & 13B   & Apache 2.0 & \url { https://huggingface.co/FreedomIntelligence/AceGPT-13B-chat}      \\
\textbf{AceGPT-v1.5-13B}        & 13B   & Apache 2.0 & \url { https://huggingface.co/FreedomIntelligence/AceGPT-v1.5-13B}      \\
\textbf{AceGPT-v1.5-13B-chat}   & 13B   & Apache 2.0 & \url { https://huggingface.co/FreedomIntelligence/AceGPT-v1.5-13B-Chat} \\
\textbf{jais-13b}               & 13B   & Apache 2.0 & \url { https://huggingface.co/core42/jais-13b}                          \\
\textbf{jais-13b-chat}          & 13B   & Apache 2.0 & \url { https://huggingface.co/core42/jais-13b-chat}                     \\
\textbf{jais-family-13b}        & 13B   & Apache 2.0 & \url { https://huggingface.co/inceptionai/jais-family-13b}              \\
\textbf{jais-family-13b-chat}   & 13B   & Apache 2.0 & \url { https://huggingface.co/inceptionai/jais-family-13b-chat} \\ \hline  
\end{tabular}
}
\end{table*}

\end{document}